\documentclass[letterpaper]{article}
\usepackage{aaai19}
\usepackage{times}
\usepackage{helvet}
\usepackage{courier}
\usepackage{url}
\usepackage{graphicx}
\title{Proppy: A System to Unmask Propaganda in Online News}
\author{Alberto Barr\'on-Cede\~no$^\dagger$  \
Giovanni Da San Martino$^\dagger$ \  Israa Jaradat$^\star$ \ Preslav Nakov$^\dagger$\\
$^\dagger$Qatar  Computing Research Institute, HBKU, Qatar\\
$^\star$University of Texas Arlington, USA\\
\{albarron, gmartino, pnakov\}@hbku.edu.qa \hspace{10mm}israa.jaradat@mavs.uta.edu\\
}
\begin{document}
\maketitle
\begin{abstract}
We present \texttt{proppy}, the first publicly available real-world, real-time propaganda detection system for online news, which aims at raising awareness, thus potentially limiting the impact of propaganda and helping fight disinformation. The system constantly monitors a number of news sources, deduplicates and clusters the news into events, and organizes the articles about an event on the basis of the likelihood that they contain propagandistic content. The system is trained on known propaganda sources using a variety of stylistic features. The evaluation results on a standard dataset show state-of-the-art results for propaganda detection.
\end{abstract}

\section{Introduction}

Propaganda is the expression of an opinion or an action by individuals or groups deliberately designed to influence the opinions or the actions of other individuals or groups with reference to predetermined ends~\cite{InstituteforPropagandaAnalysis1938}. 
We are interested in propaganda from a journalistic point of view: how news management lacking neutrality shapes information by emphasizing positive or negative aspects purposefully~\cite[p. 1]{Jowett:12}. 
Propaganda uses psychological and rhetorical techniques that are intended to go unnoticed to achieve maximum effect. 
As a result, malicious propaganda news outlets have proven to be able to achieve large-scale impact.
In particular, the power of disinformation and propaganda was arguably demonstrated during recent events, such as Brexit and the 2016 U.S. Presidential campaign.\footnote{\url{https://www.justice.gov/file/1035477/}} 

With the rise of the Web, a combination of freedom of expression and ease of publishing contents online has nurtured a number of news outlets that produce or distribute propagandistic content. Social media further amplified the problem by making it possible to reach millions of users almost instantaneously.
Thus, with the aim of helping fight the rise of propaganda, here we introduce \texttt{proppy}, a system to unmask articles with propagandistic content, which can
(\emph{i})~help investigative journalists to study propaganda online and 
(\emph{ii})~raise awareness that a news article, or a news outlet in general, might be trying to influence people's mindset.

\noindent 
To the best of our knowledge, \texttt{proppy}%
\footnote{Visit the \texttt{proppy} project at \url{http://proppy.qcri.org}}
is the first publicly available real-world, real-time monitoring and propaganda detection system for online news, which aims at raising awareness about propaganda.

\begin{figure}
\centering
 \includegraphics[scale=0.45]{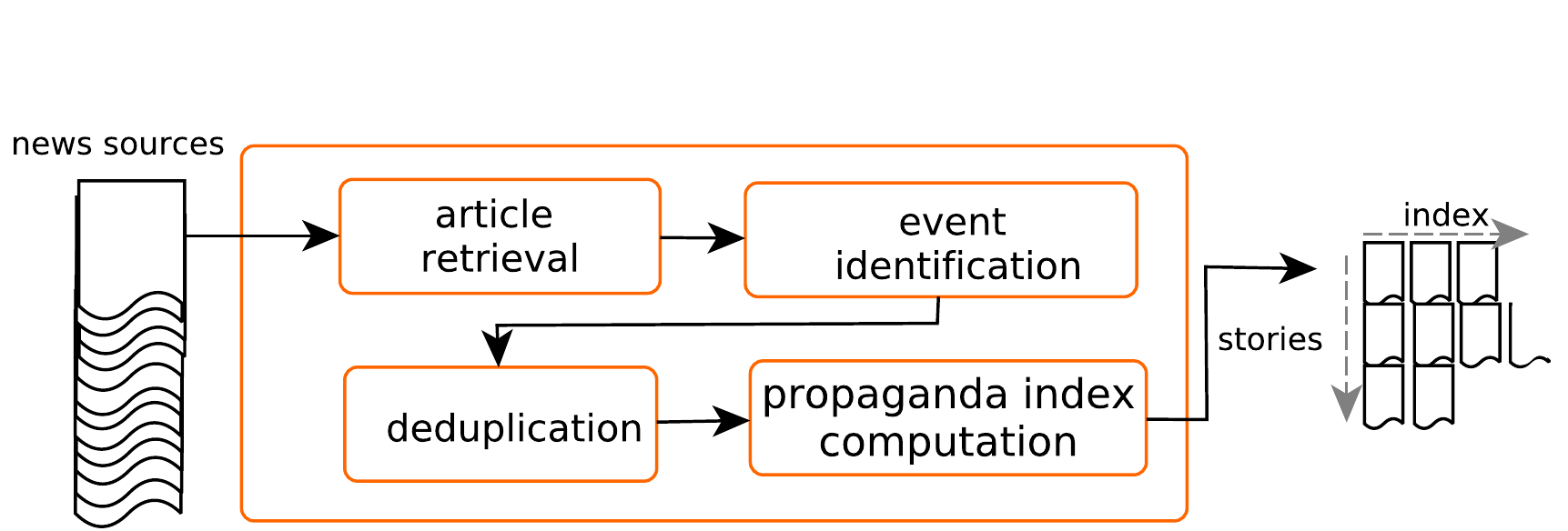}
 \caption{The architecture of \texttt{proppy}.}
 \label{fig:architecture}
\end{figure}

\section{Architecture of the System}
\label{sec:organisation}

Figure~\ref{fig:architecture} shows the architecture of \texttt{proppy}. We describe its four modules next.

\paragraph{1. Article retrieval.} 

\texttt{Proppy} regularly monitors a variety of news outlets and extracts the content of the latest news articles from their websites. 
We use GDELT\footnote{\url{http://gdeltproject.org}} to obtain links to news articles and the Newspaper3k Python library%
\footnote{\url{https://newspaper.readthedocs.io}} to extract their content.
\texttt{Proppy} then analyzes the articles in batches 
every 24 hours, and performs the remaining three steps. 

\paragraph{2. Event identification.}

We use the DBSCAN clustering algorithm~\cite{Ester:96} for event identification, as it does not require information related to the expected number of events. 
We use \textit{doc2vec} embeddings~\cite{Le:14} for article representation, pre-trained on articles from Associated Press. 
We compute the pairwise distances for DBSCAN as 1 minus the cosine similarity.
DBSCAN has two hyper-parameters: the minimum number of members in a cluster and the maximum distance between two members of the same cluster, $\epsilon$. 
We set the former parameter to 2, thus discarding singletons. 
We estimate the parameter $\epsilon=0.55$ on the METER corpus~\cite{Clough:02}. 

\paragraph{3. Deduplication.}

Next, we discard near-duplicates using a standard text re-use technique: comparison of word $n$-grams~\cite{Lyon:04} after standard pre-processing (case-folding, tokenization, and stopword removal). We compute the similarity between all pairs of documents in a cluster using the Jaccard coefficient. 
Once again, we use the METER corpus to optimize the value of $n$ and the threshold to consider two documents as near-duplicates. At run time, we discard all near-duplicates but one.

\paragraph{4. Propaganda Index Computation.\label{sec:propaganda}}

We train a maximum entropy classifier with L2 regularization to discriminate propagandistic vs non-propagandistic news articles. We use the confidence of the classifier, a value in the range $[0,1]$, to group articles into bins. 
We call this value the \textit{propaganda index}, since it reflects the probability for an article to have a propagandistic intent.
We use four families of features:

{\bf Word $n$-gram features} We use \textit{tf.idf}-weighted word $[1,3]$-grams~\cite{rashkin-EtAl:2017:EMNLP2017}.

{\bf Lexicon features.} We try to capture the \textit{typical vocabulary} of propaganda by considering representations reflecting the frequency of specific words from a number of lexicons coming from the Wiktionary, Linguistic Inquiry and Word Count (LIWC), Wilson's subjectives, Hyland hedges, and Hooper's assertives. \citeauthor{rashkin-EtAl:2017:EMNLP2017}~(\citeyear{rashkin-EtAl:2017:EMNLP2017}) showed that the words in some of these lexicons appear more frequently in propagandistic than in trustworthy articles.

{\bf Style, vocabulary richness, and readability.} Our writing style representation consists of \textit{tf.idf}-weighted character \mbox{3-grams}. This representation captures different style markers, such as prefixes, suffixes, and punctuation marks. 
We further consider the \textit{type-token ratio} (TTR) as well as the number of tokens appearing exactly once or twice in the document: \textit{hapax legomena} and \textit{dislegomena}. Moreover, we combine types, tokens, and hapax legomen\ae\ to compute Honore's R and Yule's characteristic K. We also use three readability features originally designed to estimate the level of complexity of a text: \textit{Flesch--Kincaid grade level}, \textit{Flesch reading ease} and the \textit{Gunning fog index}.

{\bf NELA.} We also integrate the NEws LAndscape (NELA) features~\cite{horne:18}: 130 content-based features collected from the existing literature that measure different aspects of a news article (e.g.,~sentiment, bias, morality, complexity).  

\begin{figure}
\centering
 \includegraphics[scale=0.2]{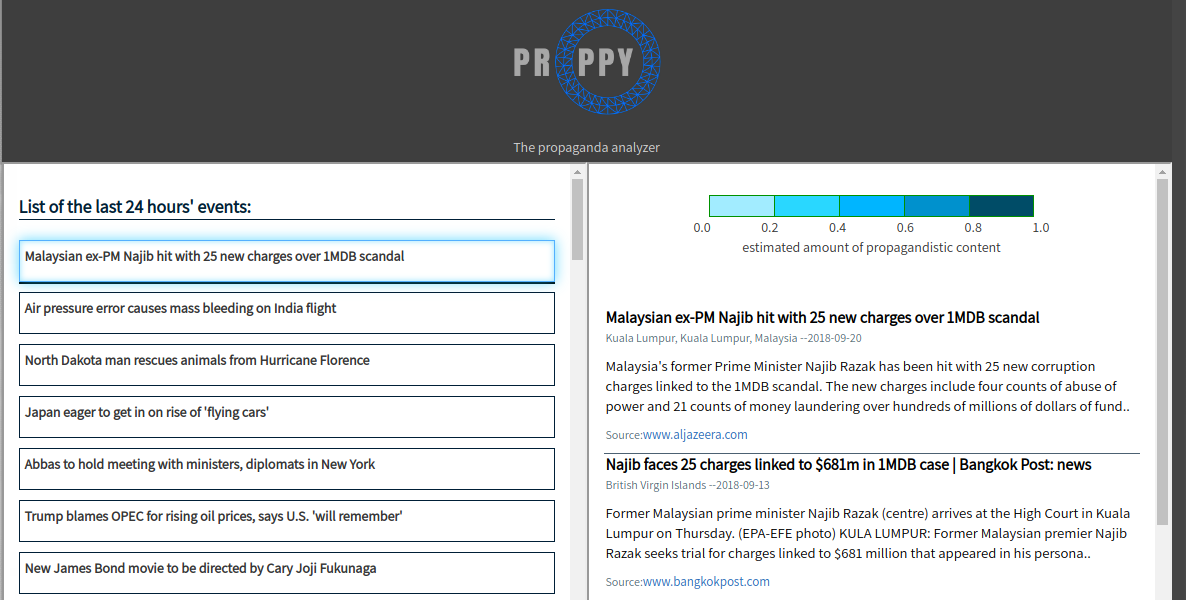}
 \caption{A screenshot of the online \texttt{proppy}.}
 \label{fig:proppy}
\end{figure}

\medskip
\noindent We evaluated \texttt{proppy} on data from~\citeauthor{rashkin-EtAl:2017:EMNLP2017}~(\citeyear{rashkin-EtAl:2017:EMNLP2017})
in a binary setup of distinguishing propaganda vs non-propaganda. 
Their $n$-grams system yielded an F$_1$ of 88.21, whereas \texttt{proppy} 
achieved 96.72 (+8.51), a statistically significant improvement (measured with the McNemar test).  

\section{Online Interface}
\label{sec:architecture}

Figure~\ref{fig:proppy} shows a screenshot of \texttt{proppy}. 
The architecture follows a \textit{push} publishing model: it updates automatically the material that it presents to the user without her taking any action but exploring the available events. 
The left panel shows events from the last 24 hours. 
When the user clicks on an event, its 
articles are shown in the right panel, organized into five bins according to their propaganda index.

\noindent The articles in bins 1 and 2 are considered nearly non-propagandistic, whereas those in the two right bins are 
propagandistic.
In this way, the user can easily observe how different media cover related events on the propaganda dimension and may guide her further exploration and judgment.

\section{Conclusion and Future Work}
\label{sec:conclusion}

We have presented \texttt{proppy}, a publicly available real-world, real-time propaganda detection system for online news, which aims at raising awareness
with the objective of limiting the impact of propaganda and helping fight disinformation. In future work, we plan to add support for multiple languages, and a \textit{pull} mode where users will be able to submit any article and get its propaganda index.

\bibliography{references}
\bibliographystyle{aaai}
\end{document}